\documentclass[conference]{csce}

\usepackage[hmargin=.75in,vmargin=1in]{geometry}
\usepackage[american]{babel}
\usepackage[T1]{fontenc}
\usepackage{times}
\usepackage{caption}

\usepackage{textcomp}
\usepackage{epsfig,graphicx}
\usepackage{xcolor}
\usepackage{amsfonts,amsmath,amssymb}
\usepackage{fixltx2e} % Fixing numbering problem when using figure/table* 
\usepackage{booktabs}
\usepackage{subfigure}
\graphicspath{ {./images/} }

\title{\bf Is preprocessing of text really worth your time for toxic comment classification?}           %%%% Replace with your title.

\author{
{\bfseries Fahim Mohammad}\\
Intel Corporations, Hillsboro, OR, USA\\     
}

\begin{document}

\maketitle                        %%%% To set Title and Author names.

\begin{abstract}%%%% Replace with your abstract.
A large proportion of online comments present on public domains are
usually constructive, however a significant proportion are toxic in nature. The
comments contain lot of typos which increases the number of features manifold, making the ML model difficult to train. 
Considering the fact that the data scientists spend approximately 80\% of their time 
 in collecting, cleaning and organizing their data \cite{CrowdFlower2016}, we explored how much effort should we 
 invest in the  preprocessing (transformation) of raw comments before feeding it to the state-of-the-art 
 classification models. With the help of four models on Jigsaw toxic comment classification data, we demonstrated 
 that the training of model without any transformation produce relatively decent model. Applying even basic
  transformations, in some cases, lead to worse performance and should be applied with caution.
\end{abstract}

\vspace{1em}
\noindent\textbf{Keywords:}
 {\small  toxic comment classification, deep learning, preprocessing, NLP, AI}
 %%%% Replace with your keywords

%%%%%%%%%%%%%%%%%%%%%%%%%%%%%%%%%%%%%%%%%%%%%%%%%%%%%%%%%%%%

\section{Introduction}
Lately, there has been enormous increase in User Generated Contents (UGC) on
the online platforms such as newsgroups, blogs, online forums and social
networking websites.  According to the January 2018 report, the number of active
users in Facebook, YouTube, WhatsApp, Facebook Messenger and WeChat was more
than 2.1, 1.5, 1.3, 1.3 and 0.98 billions respectively \cite{social_rank}. The
UGCs, most of the times, are helpful but sometimes, they are in bad taste
usually posted by trolls, spammers and bullies. According to a study by McAfee,
87\% of the teens have observed cyberbullying online \cite{mcafee2014}. The
Futures Company found that 54\% of the teens witnessed cyber bullying on social 
media platforms \cite{TheFuturesCompany2014}.  Another study found 27\% of all 
American internet users self-censor their online postings out of fear of online
harassment \cite{Lenhart2016}. Filtering toxic comments is a challenge for the
content providers as their appearances result in the loss of
subscriptions. In this paper, we will be using
\textit{toxic} and \textit{abusive} terms interchangeably to represent comments
which are inappropriate, disrespectful, threat or discriminative.

Toxic comment classification on online channels is conventionally carried out
either by moderators or with the help of text classification tools
\cite{Nobata2016}. With recent advances in Deep Learning (DL) techniques, 
researchers are exploring if DL can be used for comment classification task.
Jigsaw launched Perspective (\texttt {www.perspectiveapi.com}), which uses ML to
automatically attach a confidence score to a comment to show the extent to which
a comment is considered toxic. Kaggle also hosted an online competition on toxic
classification challenge recently \cite{kaggle_toxic}.

Text transformation is the very first step in any form of text classification. The online comments are generally in non-standard English and contain lots of spelling mistakes partly because of typos (resulting from small screens of the mobile devices) but more importantly because of the deliberate attempt to write the abusive comments in creative ways to dodge the automatic filters. In this paper we have identified 20 different atomic transformations (plus 15 sequence of transformations) to preprocess the texts. We will apply four different ML models which are considered among the best to see how much we gain by performing those transformations. The rest of the paper is organized as follows: Section 2 focuses on the relevant research in the area of toxic comment classification. Section 3 focuses on the preprocessing methods which are taken into account in this paper. Section 4 is on ML methods used. Section 5 is dedicated to results and section 6 is discussion and future work.

\section{Relevant Research }
A large number of studies have been done on comment classification in the news,
finance and similar other domains. One such study to classify comments from news
domain was done with the help of mixture of features such as the length of 
comments, uppercase and punctuation frequencies, lexical features such as  spelling, profanity and readability by applying applied linear and tree based  classifier \cite{Brand}. FastText, developed by the Facebook AI research (FAIR)  team, is a text classification tool suitable to model text involving  out-of-vocabulary (OOV) words \cite{Joulin2016} \cite{Bojanowski2006}. Zhang et  al shown that character level CNN works well for text classification without  the need for words \cite{Zhang2015}.

\subsection{Abusive/toxic comment classification}
Toxic comment classification is relatively new field and in recent years,
different studies have been carried out to automatically classify toxic
comments.Yin et.al. proposed a supervised classification method with n-grams and
manually developed regular expressions patterns to detect abusive language \cite{Yin}. Sood et. al.
used predefined blacklist words and edit distance metric to detect profanity
which allowed them to catch words such as sh!+ or @ss as profane
\cite{Sood2009}. Warner and Hirschberg detected hate speech by annotating corpus of websites and user comments geared towards detecting anti-semitic hate \cite{Warner2012}. Nobata et. al. used manually labeled online user comments from Yahoo! Finance and news website for detecting hate speech \cite{Nobata2016}. Chen et. al. performed feature engineering for classification of comments into abusive, non-abusive and undecided \cite{Chen2017}. Georgakopoulos and Plagianakos compared performance of five different classifiers namely; Word embeddings and CNN, BoW approach SVM, NB, k-Nearest Neighbor (kNN) and Linear Discriminated Analysis (LDA) and found that CNN outperform all other methods in classifying toxic comments \cite{Georgakopoulos2018}.

\subsection{Preprocessing of online comments}
We found few dedicated papers that address the effect of incorporating different
text transformations on the model accuracy for sentiment classification. Uysal
and Gunal shown the impact of transformation on text classification by taking
into account four transformations and their all possible combination on news and
email domain to observe the classification accuracy. Their experimental analyses shown that choosing appropriate combination may result in significant improvement on classification accuracy \cite{Uysal2014}. Nobata et. al. used normalization of numbers, replacing very long unknown words and repeated punctuations with the same token \cite{Nobata2016}.  Haddi et. al. explained the role of transformation in sentiment analyses and demonstrated with the help of SVM on movie review database that the accuracies improve significantly with the appropriate transformation and feature selection. They used transformation methods such as white space removal, expanding abbreviation, stemming, stop words removal and negation handling \cite{Haddi}.

Other papers focus more on modeling as compared to transformation. For example,
Wang and manning filter out anything from corpus that is not alphabet. However,
this would filter out all the numbers, symbols, Instant Messages (IM) codes, acronyms such as  \$\#!+, 13itch, </3 (broken heart), a\$\$ which gives completely different meaning to the words or miss out a lot of information. In another sentiment analyses study, Bao et. al. used five transformations namely URLs features reservation, negation transformation, repeated letters normalization, stemming and lemmatization on twitter data and applied linear classifier available in WEKA machine learning tool. They found the accuracy of the classification increases when URLs features reservation, negation transformation and repeated letters normalization are employed while decreases when stemming and lemmatization are applied \cite{Bao2014}. Jianqiang and Xiaolin also looked at the effect of transformation on five different twitter datasets in order to perform sentiment classification and found that removal of URLs, the removal of stop words and the removal of numbers have minimal effect on accuracy whereas replacing negation and expanding acronyms can improve the accuracy.

Most of the exploration regarding application of the transformation has been around the sentiment classification on twitter data which is length-restricted. The length of online comments varies and may range from a couple of words to a few paragraphs. Most of the authors used conventional ML models such as SVM, LR, RF and NB. We are expanding our candidate pool for transformations and using latest state-of-the-art models such as LR, NBSVM, XGBoost and Bidirectional LSTM model using fastText’s skipgram word vector.

\section{Preprocessing tasks}
The most intimidating challenge with the online comments data is that the words are non-standard English full of typos and spurious characters. The number of words in corpora are multi-folds because of different reasons including comments originating from mobile devices, use of acronyms, leetspeak words (\texttt{http://1337.me/}), or intentionally obfuscating words to avoid filters by inserting spurious characters, using phonemes, dropping characters etc. Having several forms of the same word result in feature explosion making it difficult for the model to train. Therefore, it seems natural to perform some transformation before feeding the data to the learning algorithm.

\begin{figure*}[tb]
\centering
\includegraphics[width=\textwidth]{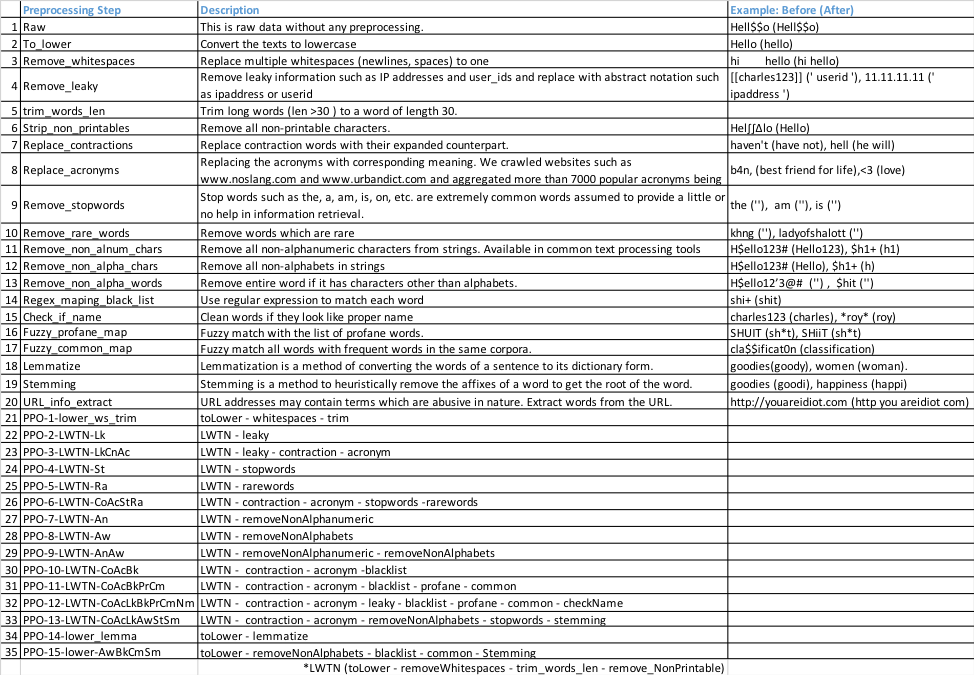} %\
\caption{List of transformations.}
\label{fig_preprocess}
\end{figure*}

To explore how helpful these transformations are, we incorporated 20 simple
transformations and 15 additional sequences of transformations in our experiment
to see their effect on different type of metrics on four different ML models
(See Figure \ref{fig_preprocess}).

\begin{itemize}
  \item Remove rare words: In the Jigsaw toxic text corpora, a staggering 65.3\%
  of the words occurred just once and 88.3\% of the words appeared five or less
  number of times (See Fig. \ref{top20}(a)). This shows that there are many
  different ways to represent the same words. The Fig. \ref{top20}(b) below
  shows different number of ways (that we could identify, actual number may be more) some of the abusive words are written in the Jigsaw corpora.
  \item Use regular expression for blacklisted words: A regular expression is created for each one of the blacklisted word and every word in corpora is compared to see which is matched. The ‘*’ (asterisk)  is assumed to be the wild character that can match any character. Our algorithm knows that s**t, S***T, sh**, shi*, s*it:), SHYT, sHYt, shiiiit, shiiiiiiiiiiiit and siht, all represent the same word.
  \item Check if the words if they look like proper name: A large number of words with frequency less than 10 looked like proper names (person, city or other proper names). We matched each words with compiled list of 1) city names 2) countries 3) nationalities 4) ethnicities 5) names of persons (a. English names, b. Spanish names, c. Hindi first names, d. Hindi last names e. Muslim names).
  \item Replace profane words using fuzzy matching: We used fuzzy matching to see how close a word is to the abusive words based on Levenshtein distance. By carefully selecting the threshold based on empirical value, the algorithm can detect that the words; SHUIT, SHYT, SHIZZ, SHiiT, SHITV, \$h1+, \$hit,  5h1t; represent the same word.
  \item Replace common words using fuzzy matching. In this transformation, we assumed that any word with a frequency of more than 100 (empirically chosen) is frequent word. Then we normalized these frequent words by removing all non-alphanumeric characters and resulted in 4,606 unique frequent words. Then, we fuzzy matched all the raw words in corpora with frequent word to get the closest word. A matching percent threshold matching\_pct is used to decide if a word is a match with a frequent word ) \begin{equation} matching\_pct = 1 - len(word)/50.\end{equation}
 
\end{itemize}

\begin{figure*}[!t]
	\centering
	\subfigure{\includegraphics[width=3.0in]{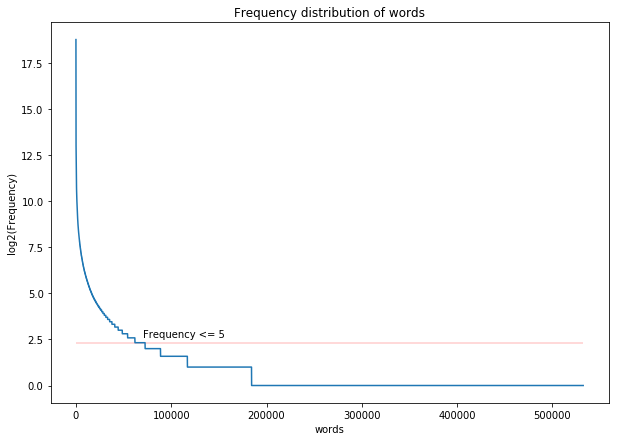}
	\label{freq_plot}}
	\hfil
	\subfigure{\includegraphics[width=3.0in]{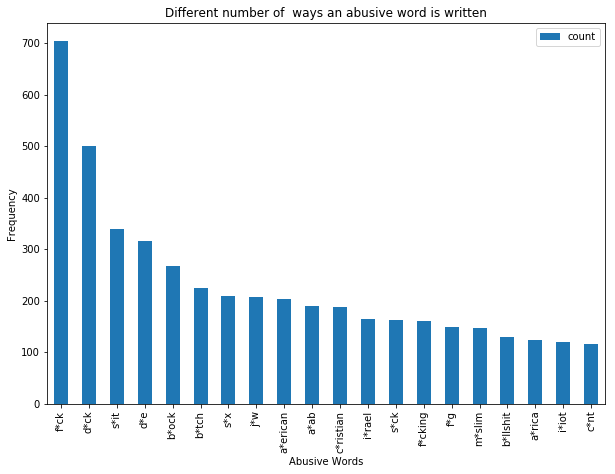}
	\label{top20}}
	\caption{a) Frequency distribution plot of the Jigsaw Toxic classification
	corpora. b) Different number of ways some of the commonly abusive words are
	written in the corpora.}
	\label{top20}
\end{figure*}

The preprocessing steps are usually performed in sequence of multiple transformations. In this work, we considered 15 combinations of the above transformations that seemed natural to us: Preprocess-order-1 through 15 in the above table represent composite transformations. For instance, PPO-11-LWTN-CoAcBkPrCm represents sequence of the following transformations of the raw text in sequence: Change to lower case $\,\to\,$ remove white spaces $\,\to\,$ trim  words len $\,\to\,$ remove Non Printable characters $\,\to\,$ replace contraction  $\,\to\,$ replace acronym  $\,\to\,$ replace blacklist using regex  $\,\to\,$ replace profane words using fuzzy  $\,\to\,$ replace common words using fuzzy.

\section{Methods}
\subsection{Datasets}
We downloaded the data for our experiment from the Kaggle’s toxic comment
classification challenge sponsored by Jigsaw (An incubator within Alphabet). The dataset contains comments from Wikipedia’s talk page edits which have been labeled by human raters for toxicity. Although there are six classes in all: ‘toxic’, ‘severe toxic’, ‘obscene’, ‘threat’, ‘insult’ and ‘identity hate’, to simplify the problem, we combined all the labels and created another label ‘abusive’. A comment is labeled in any one of the six class, then it is categorized as ‘abusive’ else the comment is considered clean or non-abusive. We only used training data for our experiment which has 159,571 labeled comments. 

\subsection{Models Used}
We used four classification algorithms: 1) Logistic regression, which is conventionally used in sentiment classification. Other three algorithms which are relatively new and has shown great results  on sentiment classification types of problems are: 2) Naïve Bayes with SVM (NBSVM), 3) Extreme Gradient Boosting (XGBoost) and 4) FastText algorithm with Bidirectional LSTM (FastText-BiLSTM). 

The linear models such as logistic regression or classifiers are used by many
researchers for Twitter comments sentiment analyses \cite{Brand} \cite{Bao2014}
\cite{Maas2011} \cite{Hamdan2015}. Naveed et. al. used logistic regression for
finding interestingness of tweet and the likelihood of a tweet being retweeted. Wang and Manning found that the logistic regression’s performance is at par with SVM for sentiment and topic classification purposes \cite{Wang2012}.

Wang and Manning, shown the variant of NB and SVM gave them the best result for sentiment classification. The NB did a good job on short texts while the SVM worked better on relatively longer texts \cite{Wang2012}. Inclusion of bigrams produced consistent gains compared to methods such as Multinomial NB, SVM and BoWSVM (Bag of Words SVM). Considering these advantages, we decided to include NBSVM in our analyses as the length of online comments vary, ranging from few words to few paragraphs. The features are generated the way it is generated for the logit model above.

Extreme Gradient Boosting (XGBoost) is a highly scalable tree-based supervised classifier \cite{Chen2016} based on gradient boosting, proposed by Friedman \cite{Friedman1999}. This boosted models are ensemble of shallow trees which are weak learners with high bias and low variance. Although boosting in general has been used by many researchers for text classification \cite{Bloehdorn2004} \cite{Kudo2004}, XGBoost implementation is relatively new and some of the winners of the ML competitions have used XGBoost \cite{Nielsen2016} in their winning solution. We set the parameters of  XGBoost as follows: number of round, evaluation metric, learning rate and maximum depth of the tree at 500,  logloss, 0.01 and 6 respectively.

FastText \cite{Bojanowski2006} is an open source library for word vector representation and text classification. It is highly memory efficient and significantly faster compared to other deep learning algorithms such as Char-CNN (days vs few seconds) and VDCNN (hours vs few seconds) and produce comparable accuracy \cite{Joulin2015}. The fastText uses both skipgram (words represented as bag of character n-grams) and continuous Bag of Words  (CBOW) method. FastText is suitable to model text involving out-of-vocabulary (OOV) or rare words more suitable for detecting obscure words in online comments \cite{Bojanowski2006}.

The Long Short Term Memory networks (LSTM) \cite{Hochreiter1997}, proposed by Hochreiter \& Schmidhuber (1997), is a variant of RNN with an additional memory output for the self-looping connections and  has the capability to remember inputs nearly 1000 time steps away. The Bidirectional LSTM (BiLSTM) is a further improvement on the LSTM where the network can see the context in either direction and can be trained using all available input information in the past and future of a specific time frame \cite{Schuster1997} \cite{Graves2004}. We will be training our BiLSTM model on FastText skipgram (FastText-BiLSTM) embedding obtained using Facebook’s fastText algorithm. Using fastText algorithm, we created embedding matrix having width 100 and used Bidirectional LSTM followd by GlobalMaxPool1D, Dropout(0.2), Dense (50, activation = ‘relu’), Dropout(0.2), Dense (1, activation = ‘sigmoid’).

\section{Results} 
We performed 10-fold cross validation by dividing the entire 159,571 comments into nearly 10 equal parts. We trained each of the four models mentioned above on nine folds and tested on the remaining tenth fold and repeated the same process for other folds as well. Eventually, we have Out-of-Fold (OOF) metrics for all 10 parts. We calculated average OOF CV metrics (accuracy, F1-score, logloss, number of misclassified samples) of all 10 folds. As the data distribution is highly skewed (16,225 out of 159,571 (~10\%) are abusive),  the accuracy metric here is for reference purpose only as predicting only the majority class every single time can get us ~90\% accuracy. The transformation, ‘Raw’, represents the actual data free from any transformation and can be considered the baseline for comparison purposes.

\begin{figure*}[tb]
\centering
\includegraphics[width=5in]{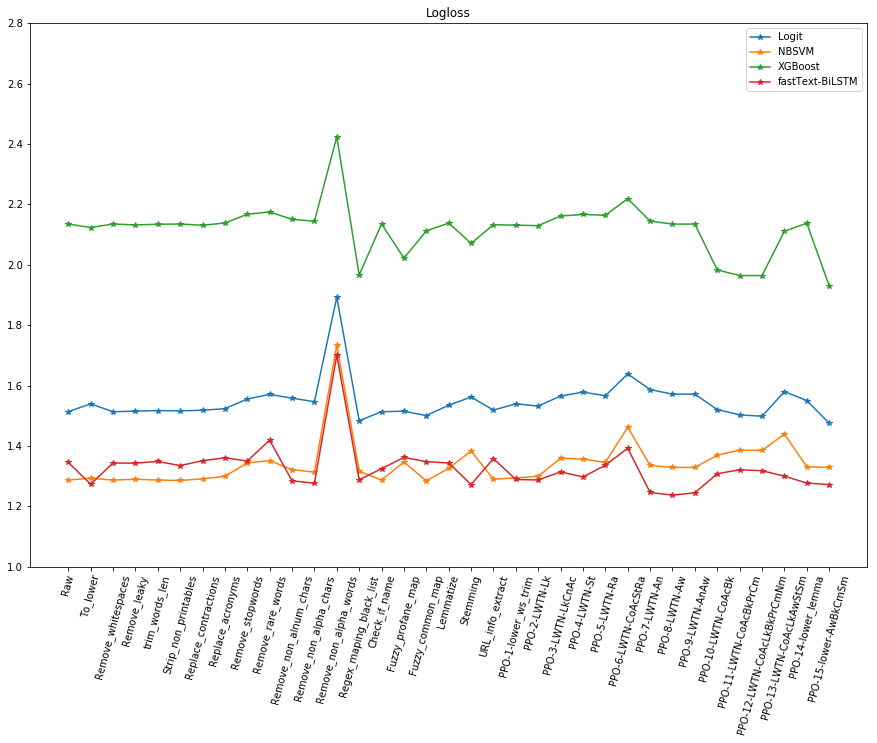}
\caption{Log loss plot for all four models on different transformations.}
\label{logloss}
\end{figure*}

\begin{figure*}[tb]
\centering
\includegraphics[width=6in]{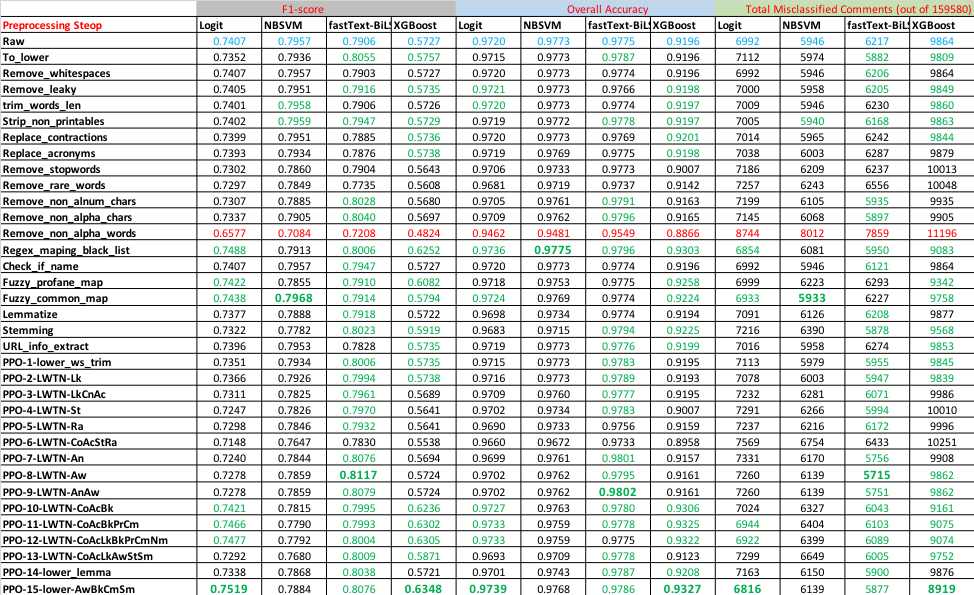}
\caption{Results: F1 scores, accuracies and total number of misclassified. }
\label{results}
\end{figure*}

Overall, the algorithms showed similar trend for all the transformations or
sequence of transformations. The NBSVM and FastText-BiLSTM showed similar
accuracy with a slight upper edge to the FastText-BiLSTM (See the logloss plot
in Fig. \ref{logloss}). For atomic transformations, NBSVM seemed to work better
than fastText-BiLSTM and for composite transformations fastText-BiLSTM was
better. Logistic regression performed better than the XGBoost algorithm and we
guess that the XGBoost might be overfitting the data. A similar trend can be
seen in the corresponding F1-score as well. One advantage about the NBSVM is
that it is blazingly fast compared to the FastText-BiLSTM. We also calculated
total number of misclassified comments (see Fig. \ref{results}).

The transformation, Convert\_to\_lower, resulted in reduced accuracy for Logit and NBSVM and higher accuracy for fastText-BiLSTM and XGBoost. Similarly, removing\_whitespaces had no effect on Logit, NBSM and XGBoost but the result of fastText-BiLSTM got worse. Only XGBoost was benefitted from replacing\_acronyms and replace\_contractions transformation. Both, remove\_stopwords and remove\_rare\_words resulted in worse performance for all four algorithms. The transformation, remove\_words\_containing\_non\_alpha leads to drop in accuracy in all the four algorithms. This step might be dropping some useful words (sh**, sh1t, hello123 etc.) from the data and resulted in the worse performance. 

The widely used transformation, Remove\_non\_alphabet\_chars (strip all non-alphabet characters from text), leads to lower performance for all except fastText-BiLSTM where the number of misclassified comments dropped from 6,229 to 5,794. The transformation Stemming seemed to be performing better compared with the Lemmatization for fastText-BiLSTM and XGBoost.

For logistic regression and the XGBoost, the best result was achieved with
PPO-15, where the number of misclassified comments reduced from 6,992 to 6,816 and from 9,864 to 8,919 respectively. For NBSVM, the best result was achieved using fuzzy\_common\_mapping (5,946 to 5,933) and for fastText-BiLSTM, the best result was with PPO-8 (6,217 to 5,715) (See Table 2). This shows that the NBSVM are not helped significantly by transformations. In contrast, transformations did help the fastText-BiLSTM significantly.

We also looked at the effect of the transformations on the precision and recall the negative class. The fastText-BiLSTM and NBSVM performed consistently well for most of the transformations compared to the Logit and XGBoost. The precision for the XGBoost was the highest and the recall was lowest among the four algorithm pointing to the fact that the negative class data is not enough for this algorithm and the algorithm parameters needs to be tuned. 

The interpretation of F1-score is different based on the how the classes are distributed. For toxic data, toxic class is more important than the clean comments as the content providers do not want toxic comments to be shown to their users. Therefore, we want the negative class comments to have high F1-scores as compared to the clean comments. We also looked at the effect of the transformations on the precision and recall of the negative class. The F1-score for negative class is somewhere around 0.8 for NBSVM and fastText-BiLSTM, for logit this value is around 0.74 and for XGBoost, the value is around 0.57. The fastText-BiLSTM and NBSVM performed consistently well for most of the transformations compared to the Logit and XGBoost. The precision for the XGBoost was the highest and the recall was lowest among the four algorithm pointing to the fact that the negative class data is not enough for this algorithm and the algorithm parameters needs to be tuned.

\section{Discussion and Future Work}
We spent quite a bit of time on transformation of the toxic data set in the hope that it will ultimately increase the accuracy of our classifiers. However, we empirically found that our intuition, to a large extent, was wrong. Most of the transformations resulted in reduced accuracy for Logit and NBSVM. We considered a total of 35 different ways to transform the data. Since, there will be exponential number of possible transformation sequences to try, we selected only 15 that we thought reasonable. Changing the order can have a different outcome as well. Most of the papers on sentiment classification, that we reviewed, resulted in better accuracy after application of some of these transformations, however, for us it was not completely true. We are not sure about the reason but out best guess is that the twitter data is character-limited while our comment data has no restriction on the size. 

The toxic data is unbalanced and we did not try to balance the classes in this experiment. It would be interesting to know what happens when we do oversampling \cite{Chawla2002} of the minority class or under-sampling of majority class or a combination of both. Pseudo-labeling \cite{Lee2013} can also be used to mitigate the class imbalance problem to some extent.

We did not tune the parameters of different algorithms presented in our experiment. It will also be interesting to use word2vec/GloVe word embedding to see how they behave during the above transformations. Since the words in these word embedding are mostly clean and without any spurious/special characters, we can't use the pre-trained word vectors on raw data. To compare apple to apple, the embedding vectors needs to be trained on the corpora from scratch which is time consuming. Also, we only considered six composite transformations which is not comprehensive in any way and will be taking this issue up in the future. We also looked only at the Jigsaw's Wikipedia data only.

This paper gives an idea to the NLP researchers on the worth of spending time on transformations of toxic data. Based on the results we have, our recommendation is not to spend too much time on the transformations rather focus on the selection of the best algorithms. All the codes, data and results can be found here: \texttt{ https://github.com/ifahim/toxic-preprocess}

\section{Acknowledgements}
We would like to thank Joseph Batz and Christine Cheng for reviewing the draft
and providing valuable feedback. We are also immensely grateful to Sasi
Kuppanagari and Phani Vadali for their continued support and encouragement
throughout this project.

%%%%%%%%%%%%%%%%%%%%%%%%%%%%%%%%%%%%%%%%%%%%%%%%%%%%%%%%%%%%
%%
%% Reference
%% Below is an example of bibliography that contains all entries within this document.
%% You can also let BibTeX generate your bibliography by inserting the following two commands:
%%
\bibliographystyle{IEEEtran}
% \bibliography{preprocess}
%% \bibliography{<your_bibliography_file_1>,<your_bibliography_file_2>,...}
%%
%% Note that you need to make sure that LaTeX (BibTeX) can find IEEEtrans.bst in your system.
%% If you are unsure about that, just place IEEEtrans.bst in the same directory where your LaTeX source files reside.
%%
%%%%%%%%%%%%%%%%%%%%%%%%%%%%%%%%%%%%%%%%%%%%%%%%%%%%%%%%%%%%%
%%% Below thebibliography environment will be automatically created in a different file (your_file_name.bbl) 
%%% if you use BibTeX and specify IEEEtrans.bst.

% \newpage
% Appendix \ref{appendix1}
% 
% \begin{figure*}[tb]
% \centering
% \includegraphics[width=\textwidth]{appendix1}
% \caption{Transformation Sequences}
% \label{appendix1}
% \end{figure*}

\end{document}